\documentclass[10pt, a4paper]{article}
\usepackage{lrec2022} % this is the new LREC2022 Style
\usepackage{multibib}
\newcites{languageresource}{Language Resources}
\usepackage{graphicx}
\usepackage{tabularx}
\usepackage{soul}
\usepackage{multirow}
\usepackage[dvipsnames]{xcolor}
\usepackage{titlesec}
\titleformat{\section}{\normalfont\large\bfseries\center}{\thesection.}{1em}{}
\titleformat{\subsection}{\normalfont\SmallTitleFont\bfseries\raggedright}{\thesubsection.}{1em}{}
\titleformat{\subsubsection}{\normalfont\normalsize\bfseries\raggedright}{\thesubsubsection.}{1em}{}
\renewcommand\thesection{\arabic{section}}
\renewcommand\thesubsection{\thesection.\arabic{subsection}}
\renewcommand\thesubsubsection{\thesubsection.\arabic{subsubsection}}
\usepackage{epstopdf}
\usepackage[utf8]{inputenc}

\usepackage{hyperref}
\usepackage{xstring}

\usepackage{color}

\title{BasqueParl: A Bilingual Corpus of Basque Parliamentary Transcriptions}

\name{Nayla Escribano\textsuperscript1, Jon Ander González\textsuperscript1, Julen Orbegozo-Terradillos\textsuperscript2,\\\normalfont\large\bfseries Ainara Larrondo-Ureta\textsuperscript2, Simón Peña-Fernández\textsuperscript2,\\\normalfont\large\bfseries Olatz Perez-de-Viñaspre\textsuperscript1 and Rodrigo Agerri\textsuperscript1}

\address{\textsuperscript1 HiTZ Center - Ixa, University of the Basque Country UPV/EHU \\
         \textsuperscript2 Gureiker, University of the Basque Country UPV/EHU \\
         \{nayla.escribano, olatz.perezdevinaspre, rodrigo.agerri\}@ehu.eus}

\abstract{Parliamentary transcripts provide a valuable resource to understand the reality and know about the most important facts that occur over time in our societies. Furthermore, the political debates captured in these transcripts facilitate research on political discourse from a computational social science perspective. In this paper we release the first version of a newly compiled corpus from Basque parliamentary transcripts. The corpus is characterized by heavy Basque-Spanish code-switching, and represents an interesting resource to study political discourse in contrasting languages such as Basque and Spanish. We enrich the corpus with metadata related to relevant attributes of the speakers and speeches (language, gender, party...) and process the text to obtain named entities and lemmas. The obtained metadata is then used to perform a detailed corpus analysis which provides interesting insights about the language use of the Basque political representatives across time, parties and gender.
 \\ \newline  }

\begin{document}

\maketitleabstract

\section{Introduction}

Parliaments and chambers of public representatives are official agoras for the presentation of ideas and their dialectical confrontation, where people chosen to represent certain groups interact in the public-political space. Minutes gathered in each of these agoras, which compile the aforementioned interactions in public institutions, can be seen as the “black box” of a given society. In this sense, we believe that speeches from public representatives may allow us to understand and interpret the reality of a given historical moment.

In this context, technological advances in Natural Language Processing (NLP), machine learning and other computational methods allow for parliamentary rhetoric to be analysed from a multidisciplinary perspective. It should be observed that public institutions have always generated large amounts of textual data (debates, laws, parliamentary discourses) that have received a special interest since the start of digitization and the emergence of the Web in the last decade of the past century. Digitization of debates allows institutions to increase their public and media impact, generating at the same time valuable textual resources that can be analyzed by computational social science and NLP research. Specifically, relevant corpora have been gathered for different NLP tasks such as sentiment analysis \cite{thomas-etal-2006-get,abercrombie-batista-navarro-2020-parlvote} or machine translation \cite{roukos1995canadian-hansard,koehn-2005-europarl,hajlaoui-etal-2014-dcep}.

With this multidisciplinary approach in mind, we gathered and processed a corpus of Basque parliamentary transcriptions for public research. Furthermore, we analyzed the representatives' speeches across different data attributes that might be of interest for the general public such as language use, gender and party. Indeed, these analyses could reflect whether political groups and concrete speakers do or do not act in parliament according to their manifested ideas. It could also serve to analyze to what extent the Basque parliament is a reflection of the society it represents. 

For a bilingual community such as the Basque Country, analyzing language use is a useful aspect to consider in this regard. Results from an official sociolinguistic poll undertaken in 2016 indicated that 13.4\% of the people in the Basque Autonomous Community spoke more Basque than Spanish, even if 33.9\% were considered to have Basque language skills\footnote{\url{https://www.irekia.euskadi.eus/uploads/attachments/9954/VI_INK_SOZLG-EH_eus.pdf}}.
%Furthermore, a poll performed by the \emph{Sociolinguistics %Cluster}\footnote{\url{https://soziolinguistika.eus/wp-content/upload%s/2021/03/a-erabilera-ing.pdf}} suggested that Basque is spoken 12.6\% of the time in Basque streets, although this study was performed not only
In this sense, our newly created corpus of parliamentary transcriptions would allow us to compare language use among the general public with respect to the linguistic behaviour of its political representatives.

In this paper we present BasqueParl, a new bilingual corpus for automatic political discourse analysis. It covers transcriptions from the Parliament of the Basque Autonomous Community for eight years and two legislative terms (2012-2020). Its main characteristic is its Basque-Spanish code-switching speeches, which have been processed to identify the language of each speech fragment. Thus, the contributions of this work include:

\begin{enumerate}
    \item The creation of BasqueParl, a new publicly available 14M word bilingual corpus for political discourse analysis.
    \item Enriching the corpus with metadata (language of each speech fragment and speaker's year of birth, gender and party) and performing neural lemmatization and NER for Basque and Spanish.
    \item A detailed data analysis showing that: (i) Basque is often used in speeches but barely to convey speech content, (ii) women are underrepresented in word production, although this trend has reversed in the last years, among other conclusions.
    \item The release of the corpus for public research\footnote{\url{https://github.com/ixa-ehu/basqueparl}}.
\end{enumerate}

We describe the Basque Parliament and relevant aspects of the legislative terms covered by the corpus in Section \ref{sec:parliament}. We discuss related work in Section \ref{sec:related}. Methods employed to process the corpus are presented in Section \ref{sec:methodology}, while the corpus processing is explained in Section \ref{sec:processing}. Finally, Section \ref{sec:analysis} shows the main results of the data analysis performed on the corpus.

% https://github.com/ixa-ehu/basque-parliament-corpus

\begin{table*}[h!]
    \centering
    \begin{tabular}{m{45em}}
        Bai, zure baimenarekin hemendik. \\
        Ba zure desioak, Guanche andrea, gureak ere badira. Harritu nau eta ez nau harritu hitza berriro hartzeak, zeren hitz egiten nengoen bitartean esan diozu albokoari \texttt{\footnotesize le voy a contestar. Le voy a contestar,} ondo iruditzen, zure eskubidean zaude, baino beno, ez dut uste inongo astakeriarik esan dudanik. \\
        Gauzak egiten dira eta uste dut nik, nik ere eskubidea dudala Gobernuak eta beste erakundeek egiten dutena esateko. Zeren beti \texttt{\footnotesize ver el vaso medio vacío o medio lleno, pues cambia un poco la perspectiva y vernos siempre en modo Gobierno, creo que no es nada objetivo. Se hacen cosas, se harán cosas y esta vez creo que me deberían reconocer que de la iniciativa primera a lo que hemos acordado, no nos hemos dejado nada o creo que casi nada. Entonces, bueno, sólo quería aclarar eso} eta eskerrak berriro. \\
        Eta ziur egon emakumea dokumentu horietan ez bada agertzen hitzetan, zeren uste dut hori ez dela garrantzitsuena, bai politiketan egongo dela eta dagoela. \\
        Eskerrik asko.
    \end{tabular}
    \caption{Example of bilingualism and code-switching in Basque Parliament speeches.}
    \label{tab:bilingualism}
\end{table*}

\section{The Basque Parliament}
\label{sec:parliament}

The term Basque Country (Euskal Herria in Basque) is usually used to refer to a number of territories across the Pyrenees between France and Spain, of which the Basque Autonomous Community (EAE) is a region.

The Basque Parliament constitutes the legislative body of the Basque Autonomous Community and controls the activity of the Basque Government. Sessions are guided and moderated by the president of the parliament, who usually uses short and frequent utterances for these purposes. Speeches can be produced in either Basque or Spanish, being many debates bilingual. Table \ref{tab:bilingualism} presents an example of a bilingual code-switching speech where the speaker changes from a language to another even inside sentences (normal font for Basque, true type for Spanish). It should be noted that the large majority of the bilingual speeches do not include as much Basque usage as this one.

Considered transcriptions include parliamentary sessions from December 3, 2012 to February 7, 2020, covering two legislative terms, namely, 2012-2016 and 2016-2020. Along these years, various political parties have taken part in debates: \emph{Euzko Alderdi Jeltzalea-Partido Nacionalista Vasco (EAJ-PNV)}, \emph{Euskal Herria Bildu (EH Bildu)}, \emph{Partido Socialista de Euskadi-Euskadiko Ezkerra (PSE-EE)}, \emph{Partido Popular (PP)}, \emph{Elkarrekin Podemos (EP)} and \emph{Unión, Progreso y Democracia (UPyD)}. However, other parties and institutions also took part, such as the ombudsman of the Basque Country (Ararteko) or \emph{Ezker Batua (EB)} by means of a regional parliamentary speaker.

EAJ-PNV is the main government party and the party of the parliament president throughout all the considered years, and is therefore the author of the largest number of speeches. UPyD had representation until 2016, and that year EP gained access to the parliament. The rest of the groups maintain their presence in parliament during all the covered time. %Table \ref{tab:speakers} shows the distribution of all the speakers that have participated along these years by party and gender.

% \begin{table*}[h!]
%     \centering
%     \begin{tabular}{l r r r r r r r r}
%         \hline\hline
%         Gender & All & EAJ-PNV & EH Bildu & PSE-EE & PP & EP & UPyD & Other \\\hline\hline
%         All & 159 & 60 & 38 & 24 & 17 & 12 & 1 & 7 \\
%         Female & 76 & 29 & 22 & 11 & 7 & 5 & 0 & 2 \\
%         Male & 82 & 31 & 16 & 13 & 10 & 7 & 1 & 4 \\
%         None/Unk. & 1 & 0 & 0 & 0 & 0 & 0 & 0 & 1 \\
%         \hline\hline
%     \end{tabular}
%     \caption{Distribution of speakers by party and gender.}
%     \label{tab:speakers}
% \end{table*}

\section{Related Work}\label{sec:related}

The analysis of parliamentary discourse has received special interest in the last years and more transcriptions and corpora are being released. Some parliaments make their transcriptions and voting data freely available for public observation, such as the UK Parliament\footnote{\url{https://www.theyworkforyou.com/}}, whereas other parliamentary sessions are released by independent sources, like GovTrack\footnote{\url{https://www.govtrack.us/}} for the US. However, transcriptions of the Basque Parliament sessions are not publicly available.

Several corpora have been built on this kind of transcripts for different purposes. There has been a special interest in tagging Sentiment Analysis on debate speeches, given the vast presence of opinions towards proposed motions. With this objective, \newcite{abercrombie-batista-navarro-2020-parlvote} present the ParlVote dataset, comprising English speeches of the UK Parliament, and perform different experiments on Sentiment Analysis. Similar approaches have been made on US congressional floor \cite{thomas-etal-2006-get}.

Parliamentary transcriptions have also been used for Machine Translation by building multilingual parallel corpora. The Canadian Hansard Corpus contains speeches from the Canadian Parliament in French and English \cite{roukos1995canadian-hansard}, whereas EuroParl \cite{koehn-2005-europarl} and DCEP \cite{hajlaoui-etal-2014-dcep} cover debates from the European Parliament nowadays in 21 and 23 European languages, respectively.

Among non-parallel multilingual corpora, ParlaMint \cite{parlamint} gathers sessions from 17 parliaments in their respective European languages (including Spanish parliament transcriptions), with transcriptions from 2015 to mid-2020 and special attention the COVID-19 period. Additionally, DutchParl \cite{marx-schuth-2010-dutchparl} covers Dutch transcriptions of parliaments from the Netherlands, Flanders and Belgium, including Dutch-French bilingual speeches from the Belgium Federal texts. This last bilingual example is perhaps the closest to our Basque-Spanish code-switching speeches which characterized our newly released corpus. In this sense, BasqueParl provides the first resource of this kind for a large language such as Spanish, the second-most native spoken language (after Chinese), and Basque, a pre-indoeuropean isolate language spoken by around 700K people.

\section{Methodology}\label{sec:methodology}

We used different systems and techniques to process and enrich the corpus with the aim of preparing it for data analysis. These resources are described below.

\subsection{Language Identification}

Since the corpus is bilingual and speeches switch from a language to another as shown in Table \ref{tab:bilingualism}, we decided to identify the language of the corpus units. To that end, we used the \emph{langdetect} language detection library\footnote{\url{https://github.com/fedelopez77/langdetect}}. \emph{Langdetect} compares the n-grams of a text to n-grams of previously built language profiles and provides the probabilities of the closest languages according to a distance metric.

\subsection{Lemmatization and Named Entity Recognition}

We performed lemmatization and Named Entity Recognition (NER) using Flair, which is both a deep learning system based on a BiLSTM architecture and a rather effective type of character-based contextual word embeddings \cite{akbik2018contextual,akbik2019flair}. The system and embeddings have demonstrated high performance in Sequence Labelling tasks such as Part-of-Speech, NER or SRL.

\newcite{Agerri2020GiveYT} demonstrate that text representation models trained on an appropriate monolingual corpus for Basque outperform large multilingual transformer-based models such as mBERT \cite{Devlin2019BERTPO} or XLM-RoBERTa \cite{Conneau2020UnsupervisedCR}. Similarly, \newcite{Agerri2020ProjectingHA} presents the Flair-Oscar monolingual model for Spanish, which obtained the best results at the CAPITEL 2020 shared task for NER in Spanish \cite{Zamorano2020OverviewOC}. These two models have been used to perform lemmatization and NER for both languages in the BasqueParl corpus.

\section{Corpus Processing}\label{sec:processing}

In order to prepare the corpus for data analysis, we performed several pre-processing steps using manual, rule-based methods and machine learning techniques. We also developed a demo to explore the pre-processed corpus.

\begin{table*}[h!]
    \centering
    \begin{tabular}{l m{35em}}
        \hline\hline
        Field & Description \\\hline\hline
        Speech date & Date corresponding to the speech, e.g. \textit{2020-02-07} \\\hline
        Speech id & Number that identifies the speech within its date, e.g. \textit{3} \\\hline
        Paragraph id & Number that identifies the paragraph within its speech, e.g. \textit{3} \\\hline
        Speaker & Family names of the speaker, including their position if any, e.g. \textit{Tejeria Otermin LEHENDAKARIA} \\\hline
        Birth & Year of birth of the speaker, e.g. \textit{1971} \\\hline
        Gender & Gender of the speaker, e.g. \textit{female} \\\hline
        Party & Political group of the speaker, e.g. \textit{EAJ-PNV} \\\hline
        Language & Language assigned to a paragraph, e.g. \textit{Basque} \\\hline
        Text & Paragraph of the speech text \\\hline
        Lemmas & Lemmatized paragraph, with and without stopwords \\\hline
        Named entities & Named entities extracted from the paragraph, with and without stopwords \\
        \hline\hline
    \end{tabular}
    \caption{Field data for each paragraph in the corpus.}
    \label{tab:fields}
\end{table*}

\begin{table*}[h!]
    \centering
    \begin{tabular}{l l r r r r r}
        \hline\hline
        Field & Category & Speeches & Paragraphs & Words & Lemmas & Entities \\
        \hline\hline
        All & All & 41,417 & 342,666 & 13,872,105 & 5,090,573 & 349,890 \\
        \hline
        \multirow{2}{4em}{Language} & Basque & 34,571 & 133,599 & 2,938,061 & 1,375,686 & 122,439 \\
        & Spanish & 18,016 & 209,067 & 10,934,044 & 3,714,887 & 227,451 \\
        \hline
        \multirow{3}{4em}{Gender} & Female & 30,857 & 185,119 & 6,452,503 & 2,424,765 & 179,030 \\
        & Male & 10,559 & 157,481 & 7,416,852 & 2,664,596 & 170,782 \\
        & None/Unknown & 1 & 66 & 2,750 & 1,212 & 78 \\
        \hline
        \multirow{8}{4em}{Party} & EAJ-PNV & 28,530 & 143,685 & 4,330,467 & 1,701,415 & 145,400 \\
        & EH Bildu & 3,638 & 58,391 & 2,423,245 & 1,009,715 & 56,440 \\
        & PP & 3,562 & 51,683 & 2,672,062 & 858,798 & 56,376 \\
        & PSE-EE & 2,834 & 44,738 & 2,327,350 & 769,767 & 43,994 \\
        & EP & 1,467 & 24,235 & 1,171,948 & 410,966 & 24,738 \\
        & UPyD & 1,368 & 19,110 & 909,007 & 322,220 & 22,270 \\
        & ARARTEKO & 10 & 627 & 30,382 & 14,535 & 524 \\
        & EB & 1 & 17 & 746 & 283 & 23 \\
        & None/Unknown & 7 & 180 & 6,898 & 2,874 & 125 \\
        \hline
        Speaker & President & 22,772 & 60,326 & 753,231 & 315,138 & 62,317 \\
        \hline\hline
    \end{tabular}
    \caption{Distribution of the corpus data.}
    \label{tab:corpus_data}
\end{table*}

\subsection{Preprocessing}
\label{ssec:preprocessing}

Basic information about each speech includes date, speech identifier, speaker, birth, gender and party. The first three fields were explicitly indicated in the original transcriptions, but we had to normalize variations (e.g. dates in different format or same person mentioned with various name forms) and correct mistakes (e.g. same speech identifier for two different speeches). In the case of gender, we built a set of rules to identify it from text. For example, \textit{andrea} (\textit{Ms.}) in the original speaker name would suggest that the speaker is a woman. On the other hand, we extracted the speaker's year of birth and party by gathering a list of politicians and their parties from official sources\footnote{\url{https://www.legebiltzarra.eus/comparla/e\_comparla\_alf\_ACT.html}}\footnote{\url{https://www.jjggalava.eus/eu/web/jjggalava/batzarkideen-historikoa}}. In exceptional cases in which some of this data was impossible to retrieve (e.g. the speaker name may be lost in the transcription process) or was not appropriate (e.g. the speech is assigned to an organization and not a person), we marked the corresponding field as none/unknown.

Additionally, we applied machine and deep learning techniques to identify language, lemmatize the text and detect named entities on originally separated speech paragraphs. We used \emph{langdetect} to identify each paragraph written in Spanish, assuming that the non-Spanish paragraphs were Basque. In cases like Table \ref{tab:bilingualism}, the system decides for each paragraph if the text is in Spanish or not. Then, we separated those paragraphs into sentences using the segtok tokenizer \footnote{\url{https://github.com/fnl/segtok}} and, according to the language detected in the paragraph, applied the corresponding Flair-based lemmatization and NER models for Basque or Spanish. Due to the agglutinative morphology of Basque, lemmatization is required to obtain named entities without their inflected forms. For both languages, we only kept named entities referring to persons, locations or organizations.

We also developed a list of stopwords for each language removing lemmas and named entities which were not of interest for our data analysis. In addition to stopwords gathered from already existing lists for Basque\footnote{\url{https://github.com/stopwords-iso/stopwords-eu/blob/master/stopwords-eu.json}} and Spanish\footnote{\url{https://github.com/6/stopwords-json/blob/master/dist/es.json}}, we filtered out lemmas appearing more than 1000 times. These lemmas mostly referred to ubiquitous terms related to the Parliament or the Government with low semantic meaning for this particular corpus.

\subsection{Quantitative description of the corpus}

Table \ref{tab:fields} illustrates the information included for each paragraph in BasqueParl. Furthermore, Table \ref{tab:corpus_data} shows the distribution of the corpus data, where words are whitespace tokenized and lemmas and entities correspond to those without stopwords. Words inherit all the field data from their paragraph, that is, if a paragraph is set to a language, all its words are also set to that language, even if a word or a sentence belongs to the other one. The same applies to lemmas and entities. All the information is distributed by language, gender and party. We also present the data for the president of the Basque Parliament, who is the author of 55\% of speeches, 18\% of paragraphs and 5\% of words.

\begin{figure}[h!]
    \centering
    \includegraphics[scale=0.37]{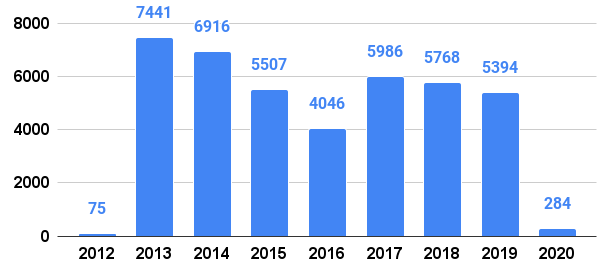}
    \caption{Number of speeches over time.}
    \label{fig:years}
\end{figure}

It should be noted that a single speech might consist of several paragraphs in both Basque and Spanish. We count a speech as belonging to either language if that language has been detected in at least one paragraph of the speech. Thus, the same speech can be counted as Basque and Spanish, which means that the sum of Basque and Spanish speeches is larger than the total number of speeches. In addition, the length of the speeches can vary significantly, from 1 paragraph to 236. These considerations will allow us to contrast language, gender and party data at speech and word level in Section \ref{sec:analysis}.

Figure \ref{fig:years} presents the distribution of speeches along the considered years. The lower numbers of speeches in 2012, 2016 and 2020 reflect changes of legislative terms and, in the case of the first and last years, the beginning and the end of the corpus.

\subsection{Demo}

We provide the BasqueParl demo\footnote{\url{http://legebiltzarra.ixa.eus}} which allows to explore the results of pre-processing and data analysis according to the fields described in Table \ref{tab:fields}, such as date, speaker, gender, party or language. Firstly, it shows speech examples and lemma and entity frequencies for the selected categories. Secondly, it provides topic modelling based on the LDA model \cite{blei2003lda}, considering documents as the non-stopword lemmas of each month. Finally, it displays scattertext plots\footnote{\url{https://github.com/JasonKessler/scattertext}} of non-stopword lemmas to compare two distinct selections of categories.

\section{Data Analysis}\label{sec:analysis}

The distribution of the corpus data described in Table \ref{tab:corpus_data} allows us to perform various analyses by crossing the information of each of the field types. This section reports the main results obtained from such analysis.

\subsection{Language}

\begin{figure}
    \centering
    \includegraphics[scale=0.37]{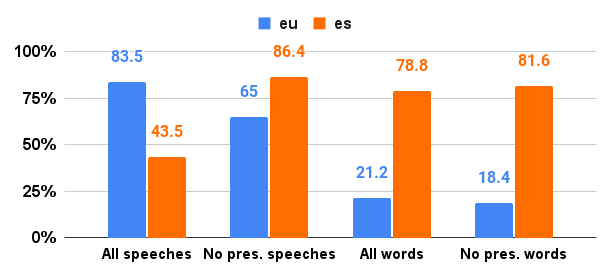}
    \caption{Language use at speech and word level.}
    \label{fig:language}
\end{figure}

An important aspect for a bilingual society is to study the language use by crossing it with other field types such as party or gender. Figure \ref{fig:language} shows the percentages of language use at speech and word level for the full corpus and ignoring the texts from the president of the parliament. As we mentioned before, a speech belongs to either language if at least one of its paragraphs belongs to that language, being possible for a speech to be in Basque and Spanish at the same time. The large amount of Basque speeches in overall (``All'') suggests that most of the speeches have at least a Basque paragraph, compared to those that have at least a Spanish paragraph. On the contrary, the number of Basque words is significantly lower than in Spanish. As observed in a  manual inspection of the corpus, this fact suggests that, even if Basque is used in most of the speeches, the most important content is usually conveyed in Spanish.

The distribution of speeches vary considerably when we filter out those texts belonging to the president of the parliament (``No pres.''), which consist mainly of short and frequent utterances in Basque like turn-takings or calls to order. Basque speeches decrease substantially while Spanish speeches double. This suggests that, although both languages are used often at least once in a speech, Spanish is more common. In contrast, this phenomenon is not reflected when we analyze language use at word level, where the percentages in language use remain independently of whether we consider the president's speeches or not. This data seems to be the most realistic and shows a gap of more than 60 percentage points between Basque and Spanish in language use.

\begin{table}[h!]
    \centering
    \begin{tabular}{c c c}
        \hline\hline
        Bilingual & Passive & None \\
        \hline\hline
        33.9\% & 19.1\% & 47.0\% \\
        \hline\hline
    \end{tabular}
    \caption{Basque language skills in EAE in 2016.}
    \label{tab:poll-knowledge}
\end{table}

% \begin{figure}[h!]
%     \centering
%     \includegraphics[scale=0.37]{figures/knowledge.png}
%     \caption{Basque language skills in EAE in 2016.}
%     \label{fig:knowledge}
% \end{figure}

\begin{figure}[h!]
    \centering
    \includegraphics[scale=0.35]{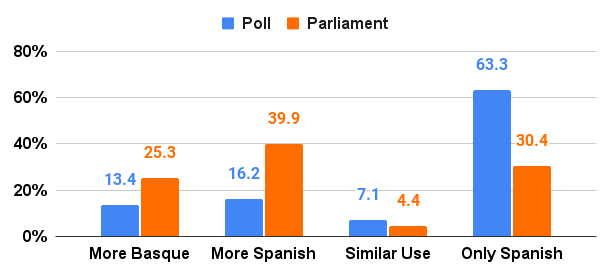}
    \caption{Language preference in EAE according to Sociolinguistic Poll 2016 and in parliament.}
    \label{fig:poll-parliament}
\end{figure}

\begin{figure*}[ht!]
    \begin{minipage}[c]{\columnwidth}
    \includegraphics[scale=0.35]{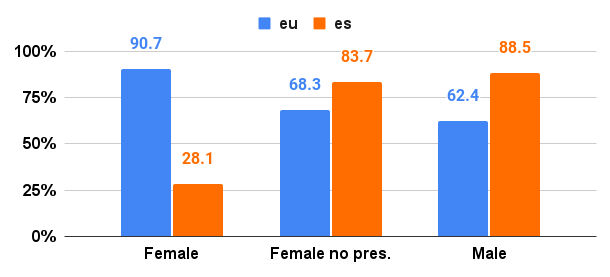}
    \caption{Language use by gender at speech level.}
    \label{fig:language-gender-spch}
    \end{minipage}
    \hfill
    \begin{minipage}[c]{\columnwidth}
    \includegraphics[scale=0.35]{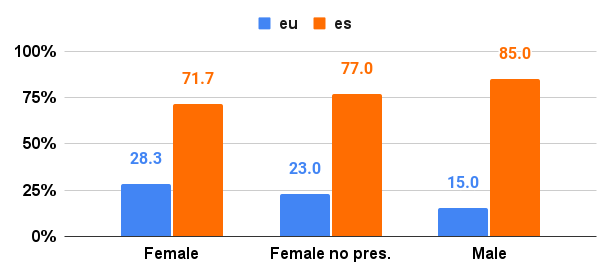}
    \caption{Language use by gender at word level.}
    \label{fig:language-gender-word}
    \end{minipage}
\end{figure*}

\begin{figure*}[ht!]
    \begin{minipage}[c]{\columnwidth}
    \includegraphics[scale=0.37]{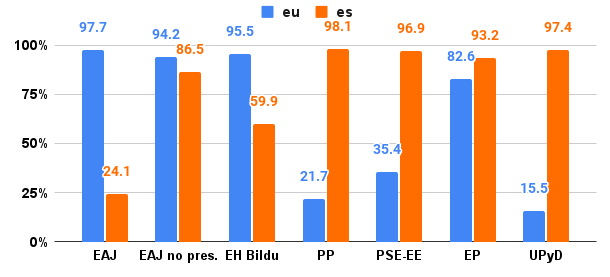}
    \caption{Language use in each party at speech level.}
    \label{fig:language-party-spch}
    \end{minipage}
    \hfill
    \begin{minipage}[c]{\columnwidth}
    \includegraphics[scale=0.37]{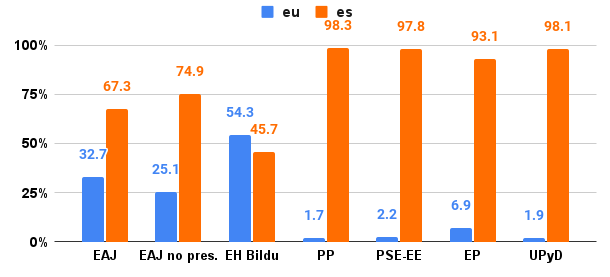}
    \caption{Language use in each party at word level.}
    \label{fig:language-party-word}
    \end{minipage}
\end{figure*}

Table \ref{tab:poll-knowledge} presents the results of Basque language skills according to the VI Sociolinguistic Poll from 2016\footnote{\url{https://www.irekia.euskadi.eus/uploads/attachments/9954/VI_INK_SOZLG-EH_eus.pdf}}, whereas Figure \ref{fig:poll-parliament} reports language preference in the Basque Autonomous Community (EAE) from the same poll and in parliament at word level. ``More Spanish'' collapses two poll categories: ``less Basque than Spanish'' and ``some Basque''. In the case of parliament results, ``More Basque" corresponds to the percentage of speakers using Basque in more than 55\% of their words, ``More Spanish" to Spanish use between 55\% and 95\%, similar use to language use between 45\% and 50\% and ``Only Spanish" to speakers using Spanish in more than 95\% of their words. It should be noted that citizens that speak Basque generally also speak Spanish. While language skills are usually higher than overall use according to the poll, the result of our analysis seems to suggest that citizens' use of Basque is lower than that of their political representatives. In fact, speakers that use more Basque than Spanish overcome citizens' result by more than 10 points. However, on a closer look, although the percentage of representatives using only Spanish is half of citizen's ratio, their preference for Spanish doubles the EAE percentage. % In this sense, and while not strictly comparable, Figure \ref{fig:poll-parliament} shows that 63.3\% of language use is done in Spanish only, whereas in parliament 78.81\% of the words are in Spanish, 81.56\% if we do not count the president's speeches. 

\subsubsection{Language by gender}

This large difference between speech level and word level in language use remains if we look at gender, as illustrated by Figures \ref{fig:language-gender-spch} and \ref{fig:language-gender-word}. Almost all speeches produced by women contain a Basque paragraph and not even a third a Spanish one, while men usually produce Spanish speeches and less often Basque ones. Again, the percentages change if we look at word numbers: the rates of language use get more similar between genders, and Spanish becomes the most frequent by more than 40 percentage points among women and 70 among men. If we ignore president's texts, female and male texts get much closer, although women still tend to use Basque more often than men.

\subsubsection{Language by party}

The data of the language use by the main political groups\footnote{We exclude EB, Ararteko and None/Unknown parties from Figures \ref{fig:language-party-spch} and \ref{fig:language-party-word} due to their limited amount of data.} in Figures \ref{fig:language-party-spch} and \ref{fig:language-party-word} reflect the same behaviour as for the overall language use. Thus, while Basque use in terms of speech is quite high (especially for EAJ-PNV, EH Bildu and EP), those rates drop wildly at word level, especially for the Spanish unionist parties (PP, PSE-EE and UPyD), for which the use of Spanish is rather non-existent. In the case of EAJ-PNV (conservative Basque Nationalist) without the president, there is also a substantial rise at the use of Spanish at speech level. In terms of words, only one party shows a larger percentage of Basque use with respect to Spanish (EH Bildu - Left Basque Pro-independence), although the numbers are quite balanced, despite contrary popular opinion. Summarizing, four parties keep Basque usage in words below 10\% and thus clearly below citizens' average usage (PP, PSE-EE, EP and UPyD), the Basque party in government EAJ-PNV conveys two thirds of the words in Spanish and only EH Bildu maintains a balanced language use.

\subsubsection{Language over time}

\begin{figure}[h!]
    \centering
    \includegraphics[scale=0.35]{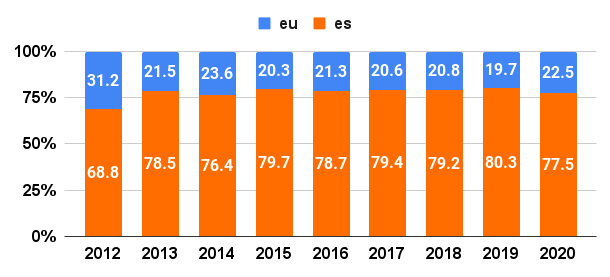}
    \caption{Language use over time at word level.}
    \label{fig:language-years}
\end{figure}

In order to check language use over time, Figure \ref{fig:language-years} shows the language use at word level across the years. Results suggest that there is no considerable change in the use of the two languages, since they keep their distance along all the considered years. However, Basque word production between 2012 and 2020 decreases almost 10 points, although these two years present too few texts and this may affect their language use. If we consider the period from 2013 to 2019, we can observe a slight reduction in Basque use.

\begin{figure*}[h!]
    \begin{minipage}[c]{\columnwidth}
    \includegraphics[scale=0.32]{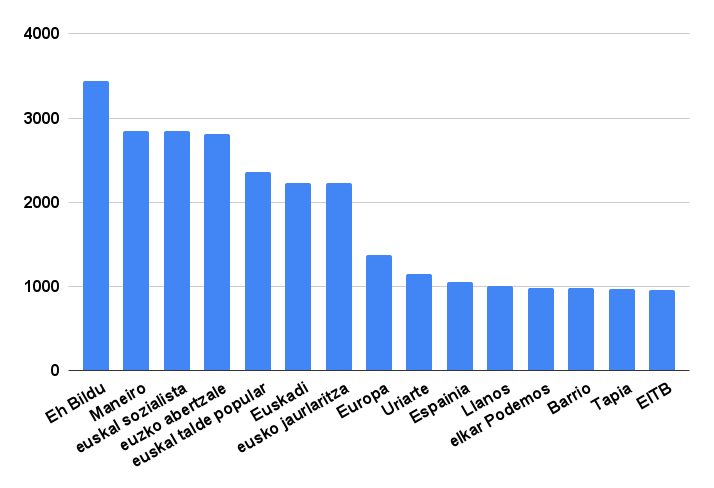}
    \caption{Most frequent Basque entities.}
    \label{fig:eu_ents}
    \end{minipage}
    \hfill
    \begin{minipage}[c]{\columnwidth}
    \includegraphics[scale=0.32]{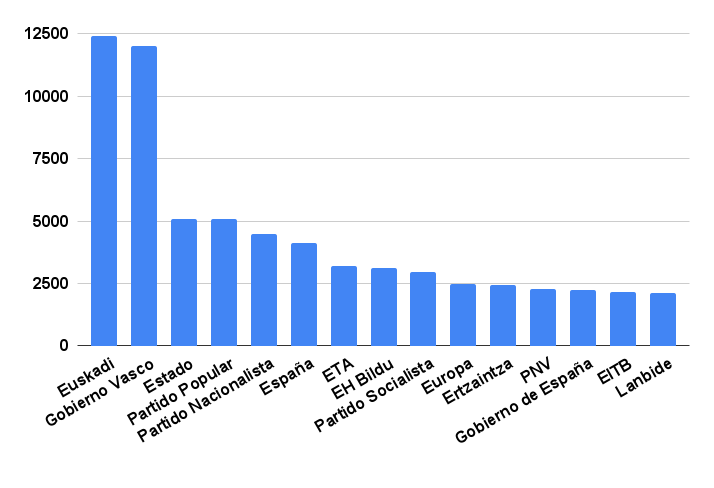}
    \caption{Most frequent Spanish entities.}
    \label{fig:es_ents}
    \end{minipage}
\end{figure*}

\begin{figure*}[h]
    \begin{minipage}[c]{\columnwidth}
    \includegraphics[scale=0.37]{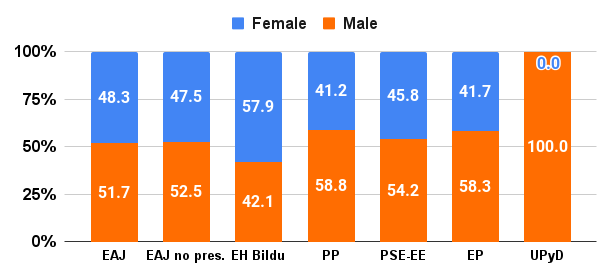}
    \caption{Gender by party at speaker level.}
    \label{fig:gender-party-spkr}
    \end{minipage}
    \hfill
    \begin{minipage}[c]{\columnwidth}
    \includegraphics[scale=0.37]{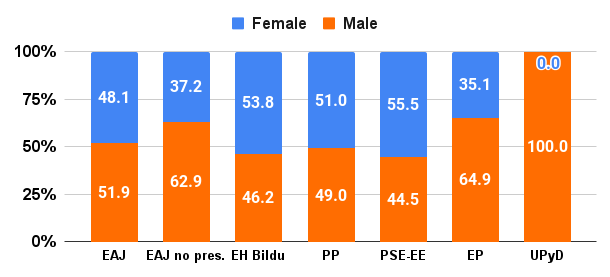}
    \caption{Gender by party at word level.}
    \label{fig:gender-party-word}
    \end{minipage}
\end{figure*}

\subsubsection{Entities by language}

Lemmatization and NER allow us to extract lemma and entity frequencies, which can serve further purposes. We present in Figures \ref{fig:eu_ents} and \ref{fig:es_ents} the most frequent entities in Basque and Spanish, respectively. As it can be seen, there are common entities of general use in parliament, like locations (e.g. \textit{Euskadi}, \textit{Espainia}/\textit{España}, \textit{Europa}), official institutions (e.g. \textit{Eusko Jaurlaritza}/\textit{Gobierno Vasco}, \textit{EITB} for public broadcast service) or political groups (e.g. \textit{EH Bildu}). However, those entities are mentioned in different frequencies: for example, \textit{Euskadi} and \textit{Gobierno Vasco} mentions double the frequency of the next entities in Spanish, whereas their Basque usage is more similar to the rest of the entities. On the other hand, Basque texts present many speaker names (e.g. \textit{Maneiro}, \textit{Uriarte}, \textit{Llanos}) and Spanish speeches add entities referred to other topics (e.g. \textit{ETA}, \textit{Ertzaintza}, \textit{Lanbide}). It must also be noted the difference in absolute frequency, being the mostly used entity in Spanish almost 4 times more frequent than the mostly used Basque entity. These results support the fact that Basque is generally used to start and end speeches and to address other speakers, while Spanish provides most of the speech content.

\subsection{Gender}

\begin{figure}[h!]
    \centering
    \includegraphics[scale=0.37]{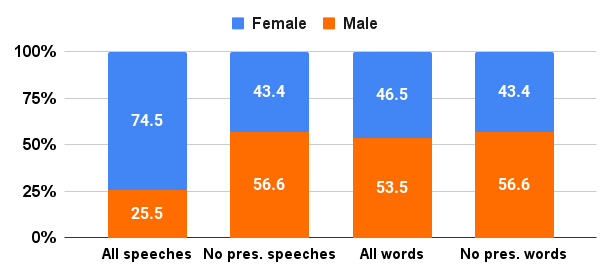}
    \caption{Gender at speech and word level.}
    \label{fig:gender}
\end{figure}

Figure \ref{fig:gender} reports percentages regarding the number of speeches or words produced by men and women. Overall, women produce most of the speeches, but less words than men. However, if we filter out the president's speeches, females drop drastically in speeches and stay below male percentages at both levels. In fact, the gap between women and men reaches more than 13 percentage points regarding speeches and words, which is three times the gap between the number of female (48\%) and male speakers (52\%) in parliament. These data would suggest that not only women speak less often, but also that they produce shorter speeches.

\subsubsection{Gender by party}

Figures \ref{fig:gender-party-spkr} and \ref{fig:gender-party-word} show speaker and word gender percentages by party for comparison.
While speaker and word level correlate in general, it is observed that two parties have more female presence at word level than the expected by the speaker rate (PSE-EE and PP), reaching almost 10 percentage points more. The rest show more male presence than perhaps expected, rising up to 10 points in the case of EAJ-PNV (without the president). In the case of UPyD, the only speaker is a man. These results indicate that, in addition to a slightly lower female representation in parliament, the rates of women interventions at speaker and word levels remain substantially lower than those of men. The only exception is EH Bildu, for which female presence is a bit higher than that of their male colleagues.

\subsubsection{Gender over time}

\begin{figure}[h!]
    \centering
    \includegraphics[scale=0.37]{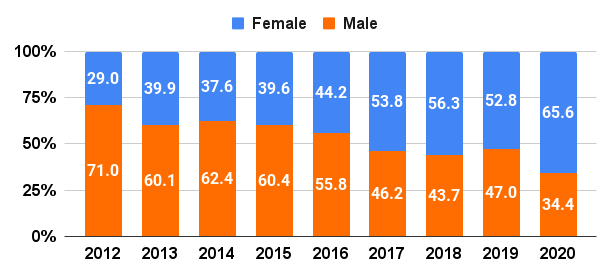}
    \caption{Gender over time at word level.}
    \label{fig:gender-years}
\end{figure}

Finally, Figure \ref{fig:gender-years} illustrate the presence of female representatives in terms of word production over time. Although in 2012 women provided less than one third of the words, by 2020 they produced almost two thirds of them. If we ignore these two years (they gather very few texts compared to the rest), there is a clear trend indicating that more female politicians are speaking more often and more at length over the passing years.

\section{Concluding Remarks}\label{sec:conclusions}

In this paper we present BasqueParl, a publicly available bilingual corpus for political discourse analysis containing Basque and Spanish transcriptions from the Basque Parliament during two legislative terms (2012-2016 and 2016-2020). The code-switching that characterizes most of the speeches offers an interesting opportunity to study language use in political debates. The transcriptions have been processed to enrich it  with metadata such as date, speaker, year of birth, gender and party. In addition, lemmas and named entities have been automatically annotated for further analysis.

The corpus data reflects relevant information about the speakers' parliamentary activity. Regarding language use, Basque is often used but barely conveys speech content, one party (EH Bildu) being the exception. If we look at gender, women participate less in parliamentary debate and, overall, their speech content is smaller than we could expect from female representation in parliament. However, this trend is being reversed in the last years.

As far as we know, BasqueParl is the only large resource (around 14M words) of its kind for Spanish and Basque. We hope that its public availability will facilitate multilingual and crosslingual research on NLP tasks related to argumentation, discourse structure, sentiment analysis and fact-checking.

\section{Acknowledgements}
This work has been partially funded by the UPV/EHU Colab 19/19 project ``Tools for the analysis of parliamentary discourses: polarization, subjectivity and affectivity in the post-truth era". Nayla Escribano is funded by the Basque Government grant ``Programa Predoctoral de Formación de Personal Investigador No Doctor del Departamento de Educación del Gobierno Vasco". Rodrigo Agerri acknowledges the support received from the RYC-2017–23647 fellowship and from the ANTIDOTE - EU CHIST-ERA project (PCI2020-120717-2) of the Agencia Estatal de Investigación through the INT-Acciones de Programación Conjunta Internacional (MINECO) 2020 call.

% \nocite{*}
\section{Bibliographical References}\label{reference}
%\label{main:ref}

\bibliographystyle{lrec2022-bib}
\bibliography{references}

% \section{Language Resource References}
% \label{lr:ref}
% \bibliographystylelanguageresource{lrec2022-bib}
% \bibliographylanguageresource{resources}

\end{document}